\documentclass[runningheads]{llncs}
\usepackage[T1]{fontenc}
\usepackage{graphicx}
\usepackage{hyperref}
\usepackage{amsfonts}
\usepackage{float}
\begin{document}
\title{Benchmarking Document Parsers on Mathematical Formula Extraction from PDFs}
\titlerunning{Benchmarking Document Parsers on Formula Extraction}
%
\author{Pius Horn\inst{1}\orcidID{0009-0004-1911-1138} \and
Janis Keuper\inst{1,2}\orcidID{0000-0002-1327-1243}}

\authorrunning{P. Horn and J. Keuper}
\institute{Institute for Machine Learning and Analytics (IMLA), Offenburg University, Offenburg, Germany
\email{pius.horn@hs-offenburg.de} \and
University of Mannheim, Mannheim, Germany}
\maketitle              
\begin{abstract}
Correctly parsing mathematical formulas from PDFs is critical for training large language models and building scientific knowledge bases from academic literature, yet existing benchmarks either exclude formulas entirely or lack semantically-aware evaluation metrics.
We introduce a benchmarking framework centered on synthetically generated PDFs with precise LaTeX ground truth, enabling systematic control over layout, formulas, and content characteristics.
For evaluation, we apply LLM-as-a-judge to assess semantic equivalence of parsed formulas, capturing mathematical meaning beyond surface-level notation differences.
We validate this approach through a human study (250 formula pairs, 750 ratings from 30 evaluators), showing a Pearson correlation of r=0.78 with human judgment, compared to r=0.34 for character-level matching (CDM) and r$\approx$0 for text similarity.
Our robust two-stage matching pipeline combining LLM-based extraction with fuzzy validation reliably aligns parsed formulas with ground truth despite format inconsistencies across parsers.
Evaluating 20+ contemporary PDF parsers across 100 synthetic documents with 2,000+ formulas reveals significant performance disparities, providing actionable guidance for practitioners selecting parsers for downstream applications.

 \keywords{PDF Document Parsing \and Mathematical Formula Extraction \and LLM-based Evaluation \and OCR Benchmarking.}
\end{abstract}
\section{Introduction}

Text extraction from PDFs is critical for training LLMs and building scientific knowledge bases.
Major training datasets rely heavily on parsed PDFs: S2ORC contains 8.1 million PDF-parsed papers (using Grobid~\cite{grobid}) versus only 1.5 million with LaTeX sources~\cite{s2orc2020}, forming the foundation of scientific components in corpora like Dolma~\cite{dolma2024}.
However, parsing quality significantly impacts downstream performance~\cite{adaparse2025}.
Creating a scientific corpus from 625K papers required over 5,000 A100 GPU hours solely for correcting PDF parsing errors~\cite{scilitllm2024}, without even evaluating formula accuracy.
Moreover, parsing limitations have left 80\% of papers from major publishers such as ACM absent from widely-used corpora like PILE and S2ORC~\cite{adaparse2025}.
Beyond LLM training, robust PDF parsing would enable better knowledge base construction, RAG systems, and semantic search~\cite{zhang2024parsing}, and could make scientific content accessible to assistive technologies given that only 3.2\% of scholarly PDFs meet accessibility standards~\cite{kumar2024accessibility}.
Despite this impact, systematic evaluation of PDF parsers for mathematical formula extraction remains understudied.

The difficulty of PDF parsing stems from fundamental design characteristics of the format.
PDFs were designed primarily for visual presentation and printing, not semantic content representation.
Most PDF documents are untagged and lack basic high-level logical structural information~\cite{chao2004layout}, making content extraction and reuse particularly challenging.
Mathematical formulas are particularly challenging due to their extensive symbol sets, two-dimensional structure where spatial positioning conveys meaning (superscripts, subscripts, fractions), and the need to convert visual arrangements into structured formats like LaTeX~\cite{zhang2024parsing}.
Consequently, no publicly available PDF parser can consistently convert documents into plain text without errors, especially for mathematical content.

To address this need, a growing number of approaches exist for converting PDFs to text~\cite{zhang2024parsing}, ranging from classic rule-based parsers to general-purpose vision-language models (VLMs) and specialized document OCR models.
These approaches differ in model size (affecting hardware requirements and runtime), accessibility (open versus closed models), and architectural design (rule-based versus vision-based).
Comparative studies have shown that parsing quality varies significantly across different document types and content characteristics~\cite{adhikari2024comparative}, with formulas and symbolic expressions posing particular challenges.
The choice of parser is therefore critical for applications such as LLM training and scientific knowledge extraction.

Despite the critical importance of parser selection for downstream applications, systematic benchmarking of PDF parsers for mathematical content extraction remains lacking.
This paper addresses this gap through the following key contributions:
\begin{itemize}
    \item We introduce a synthetic PDF generation approach with precise ground truth, overcoming limitations of manually annotated or source-derived benchmarks.
    \item We develop an LLM-based matching pipeline that reliably aligns parsed formulas with ground truth despite parser output inconsistencies.
    \item We pioneer LLM-as-a-judge for formula evaluation, demonstrating through human validation on 250 formula pairs (750 ratings from 30 evaluators) that it captures semantic equivalence more effectively than traditional metrics.
    \item We provide an open benchmark comprising two datasets: 100 synthetic PDFs with 2,000+ formulas for evaluation, and the \textit{wikipedia-latex-formulas-319k} dataset for formula generation.
    \item We establish a public leaderboard evaluating 20+ contemporary document parsers, revealing significant performance disparities and providing crucial guidance for practitioners.
\end{itemize}
Applying this framework establishes a semantically-aware, reproducible evaluation methodology for mathematical content extraction from PDFs.

\section{Related Work}
\subsection{PDF Parsing Benchmarks}
Existing PDF parsing benchmarks face systematic limitations in evaluating mathematical content.
Benchmarks derived from structured sources (e.g., \TeX{} sources from arXiv~\cite{bast2017benchmark}, XML-to-PDF matching from PubMed~\cite{publaynet,pubtabnet}) achieve large scale but explicitly exclude formulas, as mathematical content interferes with automated matching algorithms.

Manually annotated benchmarks offer broader document coverage but varying formula evaluation capabilities.
OmniDocBench~\cite{omnidocbench} (1,355 pages) evaluates display formulas using image-based character-level comparison, yet must exclude inline formulas as variability in parser output formats renders regex-based matching to ground truth unreliable.
The olmOCR-Bench~\cite{olmocr2025} employs binary unit tests requiring pixel-perfect rendering equivalence—a strict criterion that may penalize visually similar yet semantically correct formulas.
DocLayNet~\cite{doclaynet} (80,863 pages) fragments formulas into fine-grained segments unsuitable for complete extraction evaluation~\cite{adhikari2024comparative}, while the OmniAI OCR Benchmark~\cite{OmniOCRBenchmark} excludes formulas entirely.

In summary, existing benchmarks face a fundamental trade-off: structured-source benchmarks achieve scale but exclude formulas, while manually annotated benchmarks include formulas but lack semantic evaluation metrics.
Our framework addresses this gap through synthetic PDFs with precise ground truth and semantically-aware assessment.

\subsection{Formula Parsing Evaluation Metrics}
Evaluating parsed mathematical expressions is a critical challenge in document parsing.
Several datasets provide handwritten mathematical formulas paired with ground truths in LaTeX and MathML~\cite{MathBrush,ciel,CROHME,yuan2022syntax,ExpressMatch}.
While PDF formulas are machine-rendered rather than handwritten, the core challenge remains: effectively evaluating parsed formulas against ground truth.
A fundamental challenge is that LaTeX and MathML do not enforce unique representations—semantically equivalent formulas can be expressed syntactically differently.
This has motivated three main categories: text-based, tree-based, and image-based metrics.

\textbf{Text-based metrics} such as Levenshtein edit distance~\cite{levenshtein} and BLEU~\cite{bleu} have been widely adopted to assess formula parsing accuracy~\cite{cdm,omnidocbench,imagetomarkupgenerationcoarsetofineattention,wang2019translatingmathformulaimages}.

\textbf{Tree-based metrics} convert LaTeX or MathML into structural representations for comparison.
However, direct tree comparison can be biased due to this representational ambiguity.
More sophisticated structural metrics like EMERS~\cite{emers} and unbiased evaluation approaches~\cite{unbiased_evaluation} address this through graph edit distances or normalized intermediate trees, yet remain sensitive to representation ambiguity.

\textbf{Image-based approaches} render both predicted and ground truth formulas as images and compute edit distances between vertical slices of the rendered images~\cite{imagetomarkupgenerationcoarsetofineattention,wang2019translatingmathformulaimages}.
However, this approach exhibits significant limitations: a single missing character causes all subsequent positions to be flagged as mismatched, and extraneous formatting characters can produce large image differences despite semantically correct content~\cite{cdm}.
The IMEGE metric~\cite{unbiased_evaluation} performs pixel-level matching between rendered images using a bidirectional image comparison algorithm, permitting small image distortions to account for shifted sub-expressions.
While this addresses non-canonical LaTeX representations, it remains sensitive to differences in symbol sizes and visual artifacts.

Finally, CDM (Character Detection Matching)~\cite{cdm} addresses representation ambiguity through image-based character-level matching rather than direct LaTeX comparison (see Section~\ref{sec:cdm}).

\section{Methodology}
\begin{figure}
    \includegraphics[width=\textwidth]{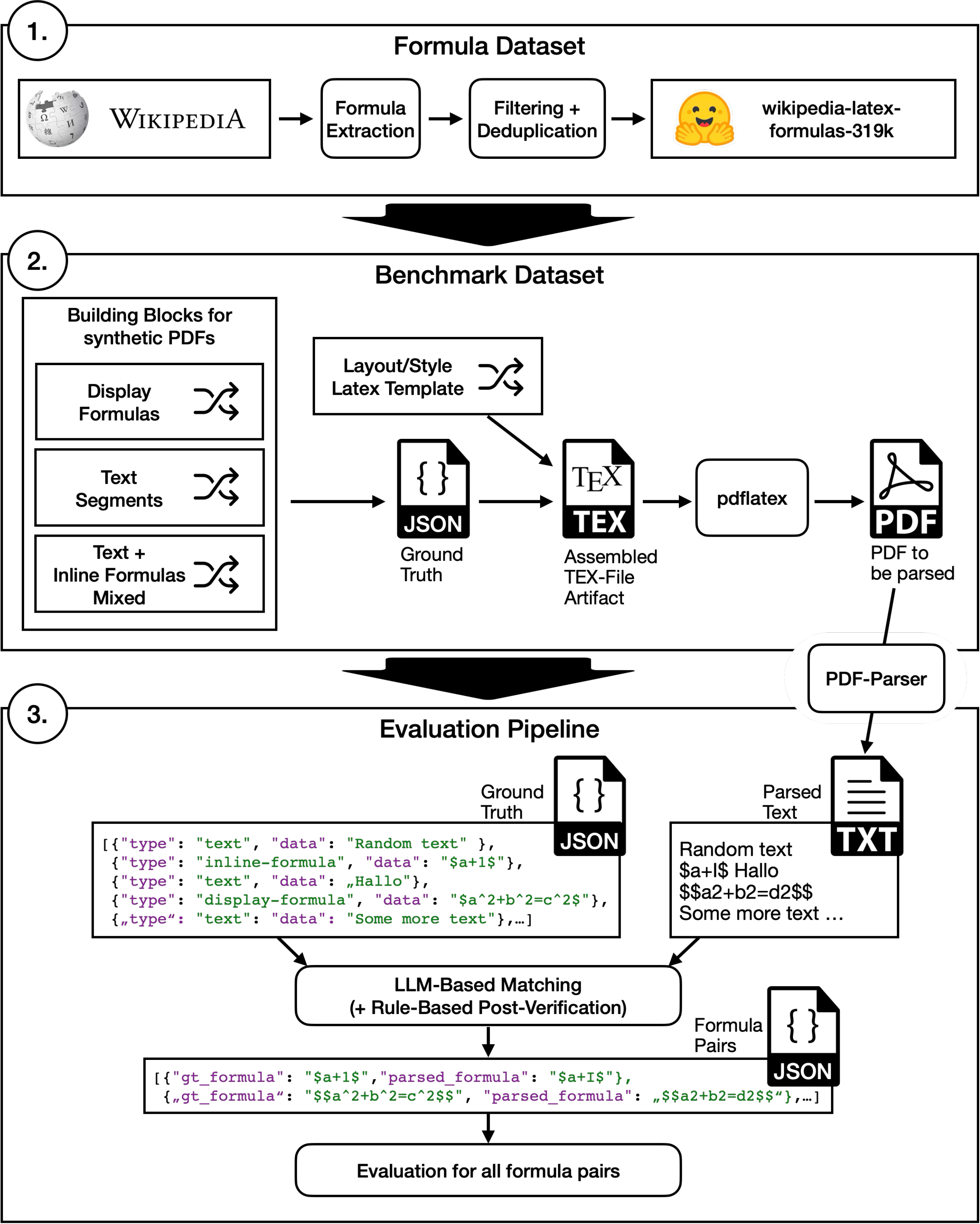}
    \caption{Overview of the three main components of the benchmarking framework.
    The formula dataset component extracts and processes mathematical formulas from Wikipedia to create the \textit{wikipedia-latex-formulas-319k} collection.
    The benchmark dataset component generates synthetic PDFs with precise ground truth by randomly combining sampled formulas from this dataset with text segments and inline formulas using randomly composed LaTeX templates.
    The evaluation pipeline component matches parsed text against ground truth using two-stage LLM-based matching and evaluates all formula pairs.} \label{fig:framework}
\end{figure}
The fundamental idea behind our parser benchmark is to generate synthetic PDFs, an approach not yet explored in the scientific literature for PDF parser evaluation.
This approach offers the distinct advantage of having the ground truth readily available.
Furthermore, it provides significant flexibility, as both the PDF's layout and its content can be precisely controlled.
This allows for the creation of PDFs tailored to ensure that parsing results can be reliably compared against the ground truth sections using a matching algorithm.
Our methodology, illustrated in Figure~\ref{fig:framework}, consists of three main components: (1)~a formula dataset extracted from Wikipedia, (2)~synthetic benchmark PDFs generated with precise ground truth, and (3)~an evaluation pipeline that matches and assesses parsed formulas.\footnote{Benchmark framework, parser implementations, and setup instructions: \url{https://github.com/phorn1/pdf-parse-bench}}

\subsection{Formula Dataset: Wikipedia Extraction}
As the benchmark aims to evaluate the capability of PDF parsers to extract formulas, the initial step involves creating a dataset composed of formulas.
We sourced this dataset from the English Wikipedia, leveraging the Wikimedia Enterprise API to access HTML dumps.
Wikipedia's use of LaTeX for rendering mathematical expressions facilitated the systematic extraction of all embedded LaTeX formulas.
To refine this collection, we filtered formulas using a visual complexity score computed by counting the total number of LaTeX commands, letters, digits, operators, brackets, punctuation marks, subscripts, and superscripts in each formula.
We retained only formulas with scores exceeding eight to exclude trivial expressions such as ``$\alpha$'', ``$x^2 + 1$'', or ``$\mathbb{R}$'', and removed duplicates.
This filtering step, which reduced the dataset by approximately half, yielded the final \textit{wikipedia-latex-formulas-319k} dataset\footnote{\url{https://huggingface.co/datasets/piushorn/wikipedia-latex-formulas-319k}}, which we contribute as part of this work and which serves as the data pool from which formulas are randomly sampled for benchmark PDF generation.

\subsection{Benchmark Dataset: Synthetic PDFs with Ground Truth}
Each benchmark PDF is generated by first creating a LaTeX template with randomized layout configurations, including document class (``article'', ``report'', ``book'', etc.), font family, page margins, font size, line spacing, paragraph indentation, and column layout (single or two-column with varying spacing).
Subsequently, we populate the template with content through an iterative compilation process designed to produce densely filled single-page documents.

At each iteration, we randomly select and append one of three content block types: (1)~plain text snippets, (2)~text snippets with embedded inline formulas, or (3)~standalone display formulas.
All formulas are randomly sampled from the \textit{wikipedia-latex-formulas-319k} dataset.
For inline formulas, we enforce a constraint that they do not exceed a height of 10 points, as taller formulas would distort line spacing and compromise visual appearance; formula heights are verified through compilation before inclusion.
Text content is generated using the Faker library\footnote{\url{https://github.com/joke2k/faker}} in one of four randomly selected languages (English, German, French, or Spanish) to introduce linguistic diversity.

After appending each content block, we compile the document with pdflatex to detect formatting issues such as content overflow or significant typesetting warnings.
If compilation reveals such issues, the most recently added block is replaced with a new randomly selected block.
This iterative process continues until adding further content would exceed the single-page limit.

\subsection{Evaluation Pipeline: Formula Matching and Metrics}

\subsubsection{Challenges in Formula Matching.}
Establishing a reliable mapping between ground truth content and parsed output presents significant challenges due to substantial variations in parser output formats.
While parsers typically delimit formulas with standard LaTeX markers such as \texttt{\$\$}, \texttt{\$}, \texttt{\textbackslash[}, \texttt{\textbackslash]}, \texttt{\textbackslash(}, or \texttt{\textbackslash)}, the consistency of this behavior varies considerably across different parsing tools.

We observed several systematic issues that complicate automated matching:
First, parsers frequently emit formulas without proper delimiters, particularly for simple expressions such as ``$a + b$'' that can be represented without explicit LaTeX syntax.
Second, multi-column layouts introduce ordering inconsistencies, as some parsers process content row-wise rather than column-wise, disrupting the expected sequential arrangement of formulas.
Third, parsers sometimes omit formulas entirely or merge multiple consecutive ground truth formulas into a single LaTeX environment in the parsed output.

These variations pose critical challenges for rule-based matching approaches such as those employed by OmniDocBench~\cite{omnidocbench}.
Such methods inevitably produce matching errors, particularly when evaluating parsers with lower output quality where formula ordering changes, formulas are omitted, or delimiters are missing.
To address these limitations, we develop a more robust matching strategy that can handle the inherent variability in parser outputs.

\subsubsection{Robust Two-Stage LLM-Based Matching}
To address the matching challenges outlined above, we developed a robust two-stage approach that combines LLM-based semantic matching with deterministic fuzzy matching validation, following the principle of turning unstable models into stable systems.

\paragraph{Stage 1: LLM-Based Formula Extraction.}
We employ GPT-5-mini to extract formulas from the parsed markdown, which we found to provide reliable extraction quality at modest computational cost.
The model receives the ordered ground truth formulas and the parsed output, and is instructed to identify each match while maintaining sequential order and extracting formulas verbatim.
The structured output returns a JSON array with extracted formulas (or empty strings if missing) and flags indicating grouped formulas that require subsequent splitting.

\paragraph{Stage 2: Fuzzy Matching Validation.}
While the LLM reliably identifies the correct formula matches, its extractions frequently differ textually from the actual parsed content due to whitespace variations and minor transcription errors.
We validate each extraction using deterministic fuzzy matching: after attempting exact substring matching, we normalize both strings by removing whitespace and backslashes, then use sliding window Levenshtein distance~\cite{levenshtein} to locate the best match in the original text.
Matches are accepted if within a threshold edit distance ratio.
For formulas that fail validation, we perform a retry with the remaining text after successful matches have been removed, allowing the LLM to focus on problematic cases.

This two-stage approach with retry mechanism combines LLM semantic understanding with deterministic precision, achieving robust formula matching despite format inconsistencies, missing delimiters, and ordering changes that would break rule-based matching.

\section{Assessment of Formula Evaluation Approaches}

Evaluating extracted formulas requires metrics that reliably compare parsed output against ground truth.
Beyond commonly considered text similarity metrics (BLEU~\cite{bleu}, Levenshtein~\cite{levenshtein}), we examine CDM~\cite{cdm}.
We omit pixel-level metrics (MSE, SSIM, IMEGE), as even slight character misalignments result in significant penalties~\cite{cdm}, and olmOCR-Bench's~\cite{olmocr2025} binary evaluation requiring identical rendering---both overly strict given that semantically equivalent formulas ($a/b$ vs. $\frac{a}{b}$, $a\,b$ vs. $ab$, $\bar{x}$ vs. $\overline{x}$) render differently yet convey identical meaning.
In addition to these established metrics, we introduce LLM-as-a-judge for semantic formula assessment.\footnote{Code, data, and human ratings for the metric study: \url{https://github.com/phorn1/formula-metric-study}}

\subsection{Human Evaluation as Reference Standard}
To establish a reliable reference for evaluating formula extraction quality, we conducted a human study in which participants assessed rendered formula pairs against ground truth for correctness, completeness, and semantic equivalence.
The evaluation dataset comprised 250 pairs of ground truth and parsed formulas (parsed using Mistral), excluding pairs where both CDM and LLM-as-a-judge assigned perfect scores, to focus on challenging cases.
Thirty participants (PhD and computer science students) each evaluated 25 pairs on a 0--10 scale, with each pair assessed by three independent raters; scores were averaged to establish robust human judgments.

Human evaluation inherently involves subjectivity, as evaluators may interpret correctness or semantic equivalence differently.
To illustrate this variability, we identified examples where raters assigned substantially different scores, demonstrating that scoring is particularly subjective for partially incorrect formulas:
\vspace{-11pt}
\begin{figure}[H]
    \includegraphics[width=\textwidth]{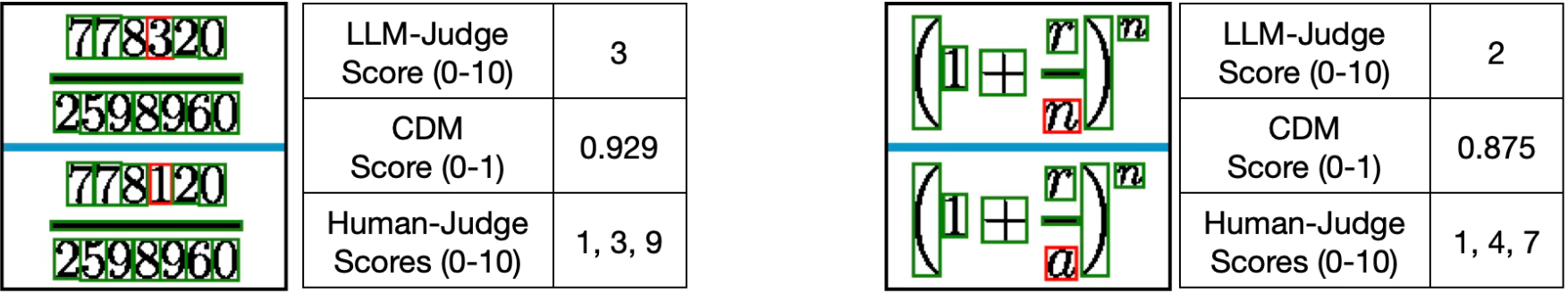}
\end{figure}
\vspace{-11pt}
All formula comparison images were generated using the official CDM tool, where green indicates correctly matched symbols and red indicates errors.

Despite this variability, we consider human evaluation a suitable reference standard for our purposes, as it captures what we prioritize in parser output: correctness, completeness, and semantic equivalence.
The aggregated human judgments thus provide a meaningful basis for validating how well automated metrics align with these criteria.

\begin{figure}
\includegraphics[width=\textwidth]{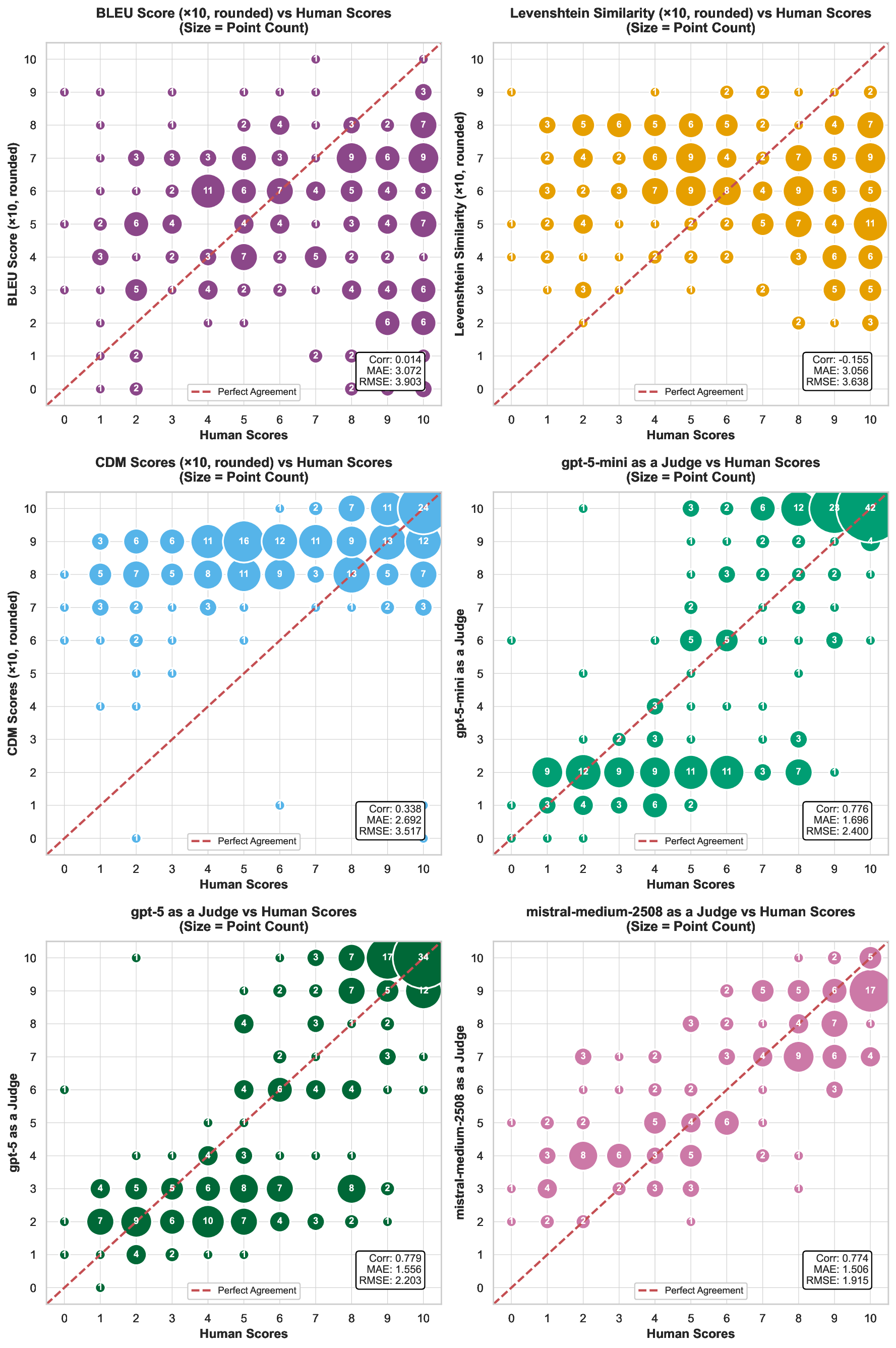}
\caption{Correlation of Automated Metrics with Human Evaluations}
\label{fig:metrics_vs_humanscore}
\end{figure}

\subsection{Text-Based Metrics: Inadequacy for Formula Evaluation}
Classical text similarity metrics (Normalized Levenshtein Distance, BLEU tokenized for \LaTeX{} codes) prove inadequate for formula evaluation due to two key issues:
(1)~\textbf{Format Variability}---parsers return formulas in \LaTeX{}, MathML, or Unicode plaintext, preventing direct comparison; and
(2)~\textbf{Representational Non-Uniqueness}---semantically equivalent formulas have multiple valid representations.
For instance, \texttt{\textbackslash frac\{1\}\{2\}}, \texttt{\textbackslash frac12}, and \texttt{\{1 \textbackslash over 2\}} all render as~$\frac{1}{2}$.
Figure~\ref{fig:metrics_vs_humanscore} confirms no correlation exists between these metrics and human scores, rendering them unsuitable for formula parsing evaluation.

\subsection{CDM: Capabilities and Systematic Limitations}
\label{sec:cdm}
CDM~\cite{cdm} addresses LaTeX representation ambiguity by rendering each formula token in a unique color, detecting character bounding boxes via image processing, and performing bipartite matching (Hungarian algorithm) based on token identity, spatial proximity, and sequential order to compute character-level precision, recall, and F1-score.

While CDM successfully identifies most incorrect or mispositioned symbols, we observe systematic limitations.
CDM produces \textit{false positives} through (1)~perfect scores despite structural errors in positional relationships (super\-scripts/\allowbreak sub\-scripts), and (2)~incorrect matching of semantically distinct symbols:
\vspace{-11pt}
\begin{figure}[H]
    \centering
    \includegraphics[width=0.7\textwidth]{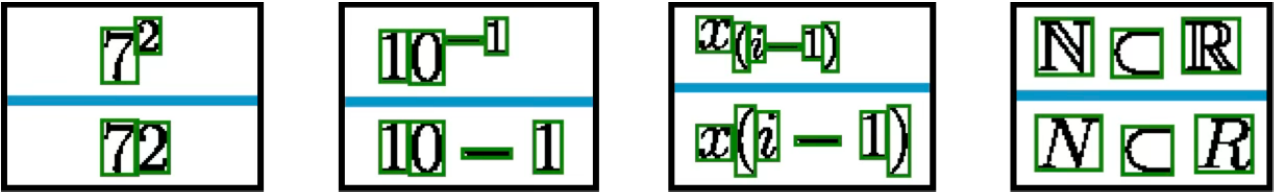}
    \label{fig:cdm_false_positives}
\end{figure}
\vspace{-11pt}
Conversely, CDM produces \textit{false negatives} in two scenarios: First, when parsers output Unicode symbols instead of LaTeX commands, which CDM cannot process.
Rule-based tools such as pypdf and pymupdf4llm output Unicode symbols exclusively, while some vision-based parsers such as dots.ocr and Nanonets-OCR-s do so frequently.
Second, though rare, CDM fails to recognize some semantically equivalent LaTeX symbol variants as identical, as shown in the examples below:
\vspace{-11pt}
\begin{figure}[H]
    \centering
    \includegraphics[width=0.7\textwidth]{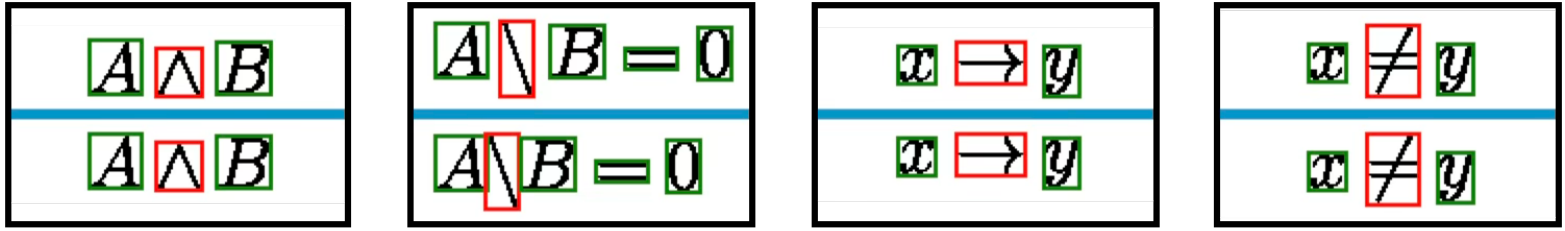}
    \label{fig:cdm_false_negatives}
\end{figure}
\vspace{-11pt}
In our human study, CDM demonstrates substantially higher correlation with human scores than simple text similarity metrics, yet exhibits notably lower correlation than LLM-as-a-judge approaches (Figure~\ref{fig:metrics_vs_humanscore}).
This discrepancy likely arises from two factors: First, CDM treats all character-level errors uniformly, whereas not all incorrect symbols are equally severe.
Second, errors that fundamentally alter a formula's semantic meaning are more strongly penalized by both human and LLM judges, while CDM's character-matching approach cannot distinguish semantically critical errors from minor notation variations.

\subsection{LLM-as-a-Judge for Semantic Formula Assessment}
We argue that the primary goal when evaluating formulas extracted by PDF parsers should be to determine whether the \emph{semantic meaning} of the original formula in the PDF has been correctly preserved.
To address the limitations of traditional metrics in capturing semantic nuance, we investigated the use of Large Language Models (LLMs) as evaluators---a technique known as ``LLM as a Judge'' that has demonstrated effectiveness across diverse domains~\cite{llm_judge}.
To the best of our knowledge, this is the first application of such an approach for formula similarity assessment in this context.

We evaluated five contemporary LLM models (GPT-5, GPT-5-mini, GPT-5-nano, Gemini-2.5-Flash, and Mistral-Medium-2508), prompting them to score formula pairs on a 0--10 scale based on correctness, completeness, and semantic equivalence.
As shown in Figure~\ref{fig:metrics_vs_humanscore}, all models exhibited substantially higher Pearson correlations with human scores (0.69--0.78) than CDM or text-based metrics, with GPT-5 and GPT-5-mini achieving the strongest correlations (0.78).
We selected GPT-5-mini as our evaluation model given its competitive performance and cost-effectiveness.
Although LLM evaluation is not infallible---occasionally assigning imperfect scores to identically rendered formulas---its correlation with human judgment (0.78) substantially exceeds that of CDM (0.34), making it a more suitable metric for evaluating mathematical formula extraction.

\section{Experiments and Results}
To systematically evaluate parser performance, we generated 100 synthetic PDF pages using our PDF generation framework, containing 1,411 inline and 641 display formulas.
We selected a broad range of parsers, including specialized OCR models such as DeepSeek-OCR~\cite{deepseek_ocr}, dots.ocr~\cite{dots_ocr}, GOT-OCR2.0~\cite{got_ocr2}, Mathpix~\cite{mathpix}, MinerU~\cite{mineru}, MonkeyOCR~\cite{monkeyocr}, Nanonets-OCR-s~\cite{nanonets_ocr_s}, Mistral OCR~\cite{mistral_ocr}, olmOCR~\cite{olmocr2025}, PaddleOCR-VL~\cite{paddleocr_vl}, and the modular PP-StructureV3 pipeline~\cite{pp_structe_v3_3}.
We also evaluated the scientific document parsers Grobid~\cite{grobid} and LlamaParse~\cite{llamaparse}, alongside general-purpose multimodal models including Gemini 2.5 Flash and Pro~\cite{gemini_2_5}, Gemini 3~\cite{gemini_3}, Qwen3-VL~\cite{qwen3_vl}, and GPT-5, GPT-5-mini, and GPT-5-nano~\cite{gpt5}.
Additionally, we included two rule-based parsers, pypdf~\cite{pypdf} and pymupdf4llm~\cite{pymupdf4llm}, which extract text from the PDF text layer and cannot process image-based documents.

Processing all 100 pages through each parser yielded over 2,000 formula pairs per parser, providing robust performance metrics.
The evaluation pipeline operates on standard CPU hardware, with total GPT-5-mini API costs below \$10 for matching and evaluating all 20+ parsers combined, ensuring broad accessibility and reproducibility.
To facilitate reproducibility and enable detailed analysis, we provide comprehensive documentation: (1)~the \textit{pdf-parse-bench} repository contains implementations and setup instructions for all evaluated parsers, including exact versions, prompts, and configuration parameters used in our benchmark; (2)~all parsing outputs and evaluation artifacts (matching results, LLM ratings) are released on Zenodo.\footnote{\url{https://doi.org/10.5281/zenodo.17806191}}
Table~\ref{tab:benchmark_results} presents the benchmark results, revealing significant performance disparities across parsers.

\begin{table}[!htb]
\centering
\caption{Benchmark results comparing parser performance on mathematical formula extraction. Score is the average LLM-as-a-Judge rating (0--10 scale) across all formulas, combining inline and display formulas. Separate scores for inline/display formulas and CDM metrics are available on the project GitHub page. Params/Cost shows active parameters for open-source models or API pricing for commercial services (as of December 2025); note that pricing models vary (per token vs. per page). Inference shows deployment options.}
\label{tab:benchmark_results}
\small
\begin{tabular}{@{}rlclc@{}}
\hline
\textbf{Rank} & \textbf{Parser} & \textbf{Score} & \textbf{Params/Cost} & \textbf{Inference} \\
\hline
1 & Qwen3-VL-235B-A22B-Instruct~\cite{qwen3_vl} & 9.76 & 22B & GPU/API \\
2 & Gemini 3 Pro~\cite{gemini_3} & 9.75 & \$2.00/12.00 M tok & API \\
3 & PaddleOCR-VL~\cite{paddleocr_vl} & 9.65 & 0.9B & CPU/GPU \\
4 & Mathpix~\cite{mathpix} & 9.64 & \$0.005/page & API \\
5 & dots.ocr~\cite{dots_ocr} & 9.43 & 1.7B & GPU \\
6 & PP-StructureV3~\cite{pp_structe_v3_3} & 9.34 & <0.3B & CPU/GPU \\
7 & Nanonets-OCR-s~\cite{nanonets_ocr_s} & 9.31 & 4B & GPU \\
8 & Gemini 2.5 Pro~\cite{gemini_2_5} & 9.28 & \$1.25/10.00 M tok & API \\
9 & MonkeyOCR-pro-3B~\cite{monkeyocr} & 9.25 & 3B & GPU \\
10 & MinerU2.5~\cite{mineru} & 9.17 & 1.2B & CPU/GPU/API \\
11 & olmOCR-2-7B-1025-FP8~\cite{olmocr2025} & 8.94 & 7B & GPU \\
12 & Gemini 2.5 Flash~\cite{gemini_2_5} & 8.78 & \$0.15/0.60 M tok & API \\
13 & Mistral OCR~\cite{mistral_ocr} & 8.66 & \$0.001/page & API \\
14 & DeepSeek-OCR~\cite{deepseek_ocr} & 8.55 & 0.6B & GPU \\
15 & LlamaParse~\cite{llamaparse} & 8.14 & \$0.09/page & API \\
16 & GPT-5 nano~\cite{gpt5} & 7.79 & \$0.05/0.40 M tok & API \\
17 & PyPDF~\cite{pypdf} & 7.69 & —/Free & CPU \\
18 & GOT-OCR2.0~\cite{got_ocr2} & 7.38 & 0.58B & CPU/GPU \\
19 & PyMuPDF4LLM~\cite{pymupdf4llm} & 6.67 & —/Free & CPU \\
20 & GPT-5 mini~\cite{gpt5} & 6.61 & \$0.25/2.00 M tok & API \\
21 & GROBID~\cite{grobid} & 5.70 & —/Free & CPU \\
\hline
\end{tabular}
\end{table}

\section{Discussion}
The benchmark reveals significant performance disparities: state-of-the-art vision-language models (Qwen3-VL, Gemini 3 Pro) and specialized document OCR systems (PaddleOCR-VL, Mathpix) achieve scores exceeding 9.6, while traditional rule-based parsers and general-purpose VLMs without document specialization perform considerably weaker.
The leaderboard is dominated by recent 2025 models, suggesting rapid progress in this domain.
However, even parsers with high overall scores occasionally produce severe parsing errors, though such failures become increasingly rare among top-performing systems.

Our evaluation methodology demonstrates that semantic assessment captures formula quality more effectively than traditional metrics.
The strong correlation between LLM-as-a-judge and human judgment validates this approach as suitable for evaluating mathematical content extraction, addressing the limitations of character-level matching which treats all errors uniformly without considering semantic impact.

\textbf{Limitations.}
Synthetic PDFs may not capture all real-world document diversity (scanned documents, historical papers, publisher-specific artifacts), and our evaluation focuses exclusively on formulas without assessing text, tables, or complex layouts.
While LLM-as-a-judge substantially outperforms traditional metrics, it is not infallible, occasionally assigning imperfect scores to correct formulas or missing subtle semantic errors.
Additionally, it requires proprietary models, though evaluation costs remain minimal.

\textbf{Future Work.}
The synthetic generation and LLM-based evaluation methodology provides a scalable foundation for extending evaluation to additional document elements (tables, figures, etc.) while maintaining precise ground truth control and semantic-aware assessment.

\subsubsection{Acknowledgements} This work has been supported by the German Federal Ministry of Research, Technology, and Space (BMFTR) in the program ``For\-schung an Fach\-hoch\-schu\-len in Ko\-ope\-ra\-tion mit Un\-ter\-neh\-men (FH-Koope\-rativ)'' within the joint project \textit{LLMpraxis} under grant 13FH622KX2.

\bibliographystyle{splncs04}
\bibliography{references}

\end{document}